\documentclass[letterpaper, 10 pt, conference]{ieeeconf}  

\IEEEoverridecommandlockouts                              

\usepackage{url}

\usepackage{cite}                      
\usepackage{tabu}                      
\usepackage{booktabs}                  

\usepackage{algorithm}

\usepackage{graphics} 
\usepackage{epsfig} 
\usepackage{amsmath}
\usepackage{amssymb}  
\usepackage{xspace}
\usepackage{overpic}
\usepackage{bm}

\usepackage{hyperref}
\usepackage{url}
\usepackage{graphicx}
\usepackage{subfigure}
\usepackage{color}
\usepackage{float}
\usepackage{textcomp}
\usepackage{multirow}
\usepackage{algpseudocode}

\newcommand{\tabincell}[2]{\begin{tabular}{@{}#1@{}}#2\end{tabular}}

\title{\LARGE \bf
Transferring Grasp Configurations using Active \\Learning and Local Replanning
}

\author{
Hao Tian$^{1, 2}$, Changbo Wang$^{1}$, Dinesh Manocha$^{2}$ and Xinyu Zhang$^{1}$
\thanks{$^{1}$School of Computer Science and Software Engineering, East China Normal University, Shanghai, China.}
\thanks{$^{2}$Department of Computer Science and Electrical $\&$ Computer Engineering, University of Maryland at College Park, MD, USA.}%
}

\begin{document}

\maketitle
\thispagestyle{empty}
\pagestyle{empty}

\begin{abstract}
We present a new approach to transfer grasp configurations from prior example objects to novel objects. We assume the novel and example objects have the same topology and similar shapes. We perform 3D segmentation on these objects using geometric and semantic shape characteristics. We compute
a grasp space for each part of the example object using active learning. We build bijective contact mapping between these model parts and compute the corresponding grasps for novel objects. Finally, we assemble the individual parts and use local replanning to adjust grasp configurations while maintaining its stability and physical constraints.
Our approach is general, can handle all kind of objects represented using mesh or point cloud and a variety of robotic hands.


\end{abstract} 
\section{Introduction}
Robot grasping has a wide range of applications in industrial automation and humanoid service robots. 
Given an object, the goal of grasping is first to compute feasible and stable grasps and then to execute grasping tasks using a gripper or a multi-fingered robotic hand. Many techniques have been proposed to interpret, compute, or evaluate grasps and their stability.
These techniques  can be broadly categorized into physically-based approaches~\cite{Bicchi2000grasping,markenscoff1989optimum,fischer1997fast,elkhoury2009on} and data-driven approaches~\cite{bohg2014data,pelossof2004an,Diankov2010Automated,goldfeder2009columbia,Zacharias2009}.
Recently, machine learning techniques have also been used for planning robust grasps~\cite{levine2018learning,mahler2016dex,mahler2017dex}.

Some data-driven approaches aim to construct a grasp database for various models by computing a set of stable grasp configurations for each object. However, it is very computationally expensive to compute stable grasp spaces when many new objects are added to the grasp database. Therefore, reusing the grasps of similar shapes is promising ~\cite{li2007data,goldfeder2009columbia,brook2011collaborative,hillenbrand2012transferring,vahrenkamp2016part,stouraitis2015functional}, as opposed to computing stable grasps for the novel objects from scratch.
These works mimic the way that humans learn grasping behaviors based on object categories that group objects with similar topologies and geometric shapes~\cite{aleotti2011part}. 
Moreover, studies in neuro-psychology for object manipulation indicate that, when humans perceive an object to grasp, the object is parsed into a few constituent parts with different affordances~\cite{biederman1987recognition,hoffman1984parts}. Some parts of an object are designed to be suitable for grasping.
For instance, a handle of a mug is designed for grasping.
Such a parsing process corresponds to segmenting an object into different semantic and functional parts.
To perform good grasps, the key is to find a suitable part of the given object for grasping.

In this paper, we present a new approach to transfer grasp configurations from prior example objects to novel objects. The objects are first segmented into the same semantic parts. We compute
a grasp space for each part of the example object and transfer it to the corresponding part of the novel objects using bijective contact mapping. Finally, we assemble the individual parts of the novel object and adjust their associated grasps using local replanning.
Here, we present a summary of our contributions.
\begin{itemize}
  \item We sample configuration space and compute grasp spaces for example objects using active learning and particle swarm optimization. This allows efficiently searching for potential grasps in high dimensional configuration space.
  \item We propose a new hybrid grasp measure for determining stable grasp configurations. It takes account of both hand-object shape fitting and  grasp quality defined in wrench space.
  \item Grasp transfer through bijective contact mapping not only computes a new grasp that is similar to the original grasp, but it can also adjust grasp to ensure the stability and physical constraints of the overall grasp of the novel objects.
  \item In local replanning for novel objects, we define a new objective function, accounting for contact points, normals, joint angles, and force closure-based grasp quality. Local replanning is also used to avoid collision with other parts and ensure stability.
%
%
  \item Our algorithm is very general in the sense that it can deal with polygon meshes and point clouds, and can be applied to high-DOF dexterous hands. This part-based grasps can be used to perform task-specific grasping.
\end{itemize}



In our benchmarks, we use a three-fingered Barrett hand to test our algorithm on three categories of models, including non-zero-genus or complex objects.
{Our method has a high success rate of learning grasp configurations for novel objects, ranging from $72.5\%$ to $92.5\%$ for different objects. Our method can achieve up to $52.6\%$ improvement in success rate as compared to a prior method~\cite{krug2014optimization}.}

\section{Related Work}\label{sec:related}
In this section, we give a brief overview of prior work on object grasping, especially grasp transfer and part-based grasping.

\emph{Grasping Similar Objects:}
These techniques rely on the fact that objects can be grouped into categories with common characteristics, such as usage, application, or geometric shape. Such categories must be known for grasping tasks. Nikandrova et al.~\cite{nikandrova2015category} demonstrated a category-based approach for object grasping. Different methods have been proposed to determine the object categories automatically~\cite{marton2009probabilistic,madry2012object} and define an object representation and a similarity metric for grasp transfer~\cite{bohg2014data}.
For a novel object, a known similar object and its preplanned grasps were retrieved from a grasp database~\cite{goldfeder2009columbia,li2007data}.

\emph{Grasp Transfer:}
Shape similarity has been used to transfer grasps to novel objects~\cite{li2007data,goldfeder2009columbia,brook2011collaborative}.
Vahrenkamp et al.~\cite{vahrenkamp2016part} transferred grasp poses according to their shape and local volumetric information. 
Grasp transfer was used to preserve the functionality of pre-planned grasps using surface warping~\cite{hillenbrand2012transferring}. In~\cite{stouraitis2015functional,amor2012generalization}, grasp poses were transferred using a contact warp method suggested in~\cite{hillenbrand2010non}. This method minimized the distance between the assigned correspondences. The warped contacts were locally replanned to ensure grasp stability. Diego et al.~\cite{rodriguez2018transferring-skill} transferred manipulation skills to novel objects using a non-rigid registration method. This work was extended by accumulating grasping knowledge and shape information~\cite{rodriguez2018transferring-category}.

\emph{Part-based Grasping:}
Many techniques have been proposed to segment objects into parts and perform grasp planning on the resulting parts. In~\cite{goldfeder2007grasp,huebner2008minimum}, objects were represented with simplified data structures such as superquadric and minimum volume bounding boxes to reduce the complexity of planning grasps.
Aleotti et al.~\cite{aleotti2011part} proposed an approach based on programming by demonstration and 3D shape segmentation. Their shape segmentation was based on Reeb graphs that were used to generate a topological representation of the objects.

\newcommand{\A}{\mbox{$\mathbf A$}\xspace}
\newcommand{\Ad}{\mbox{$\mathbf A_{\delta}$}\xspace}
\newcommand{\Ax}{\mbox{$\mathbf {A}_{\mathbf{x}}$}\xspace}
\newcommand{\B}{\mbox{$\mathbf B$}\xspace}
\newcommand{\cc}{\mbox{$\mathbf c$}\xspace}
\newcommand{\pp}{\mbox{$\mathbf p$}\xspace}
\newcommand{\np}{\mbox{$\mathbf{n}^\mathbf{p}$}\xspace}
\newcommand{\npi}{\mbox{$\mathbf{n}_i^\mathbf{p}$}\xspace}
\newcommand{\ci}{\mbox{$\mathbf c_i$}\xspace}
\newcommand{\nc}{\mbox{$\mathbf {n^c}$}\xspace}
\newcommand{\nci}{\mbox{$\mathbf{n}_i^\mathbf{c}$}\xspace}
\newcommand{\Bd}{\mbox{$\mathbf B_{\delta}$}\xspace}
\newcommand{\By}{\mbox{$\mathbf {B}_{\mathbf{y}}$}\xspace}
\newcommand{\bigTheta}{\mbox{$\mathbf {\Theta}$}\xspace}
\newcommand{\ap}{\mbox{$\mathbf a$}\xspace}
\newcommand{\ai}{\mbox{$\mathbf a_i$}\xspace}
\newcommand{\aphat}{\mbox{$\mathbf a$}\xspace}
\newcommand{\bp}{\mbox{$\mathbf b$}\xspace}

\newcommand{\bj}{\mbox{$\mathbf b_j$}\xspace}
\newcommand{\ba}{\mbox{$\mathbf {b_a}$}\xspace}
\newcommand{\bai}{\mbox{$\mathbf {b}_{\mathbf{a}_i}$}\xspace}
\newcommand{\bahat}{\mbox{$\mathbf {b_a}$}\xspace}
\newcommand{\dw}{\mbox{$\mathbf d$}\xspace}
\newcommand{\ndwi}{\mbox{$\mathbf{n}_i^\mathbf{d}$}\xspace}
\newcommand{\ndwj}{\mbox{$\mathbf{n}_j^\mathbf{d}$}\xspace}
\newcommand{\nii}{\mbox{$\mathbf n_i$}\xspace}
\newcommand{\na}{\mbox{$\mathbf {n^{\ap}}$}\xspace}
\newcommand{\nai}{\mbox{$\mathbf n^{\ap_{i}}$}\xspace}
\newcommand{\nba}{\mbox{$\mathbf n^{\ba}$}\xspace}

\newcommand{\nb}{\mbox{$\mathbf {n^b}$}\xspace}
\newcommand{\nbi}{\mbox{$\mathbf{n}_i^\mathbf{b}$}\xspace}
\newcommand{\nbj}{\mbox{$\mathbf{n}_j^\mathbf{b}$}\xspace}
\newcommand{\RR}{\mbox{$\mathbf R$}\xspace}
\newcommand{\TT}{\mbox{$\mathbf T$}\xspace}
\newcommand{\Rspace}{\mbox{$\mathbb R$}\xspace}
\newcommand{\Tspace}{\mbox{$\mathbb R^3$}\xspace}
\newcommand{\f}{\mbox{$f$}\xspace}
\newcommand{\F}{\mbox{$F$}\xspace}
\newcommand{\Back}{\mbox{$F^{-1}$}\xspace}
\newcommand{\xp}{\mbox{$\mathbf x$}\xspace}
\newcommand{\yp}{\mbox{$\mathbf y$}\xspace}
\newcommand{\FF}{\mbox{$\textsl F$}\xspace}
\newcommand{\rr}{\mbox{$\mathbf r$}\xspace}

\section{Overview}\label{sec:overview}

\subsection{Problem Definition and Notations}
Given an example object and a novel object, our goal is first to compute the grasp space for the segmented example object and transfer that knowledge to a novel object.
We assume the novel object and the example object belong to the same category in which objects share the same topologies and have similar shapes~\cite{vahrenkamp2016part,rodriguez2018transferring-category}.

\begin{figure}[htb]
\centering
\epsfig{file=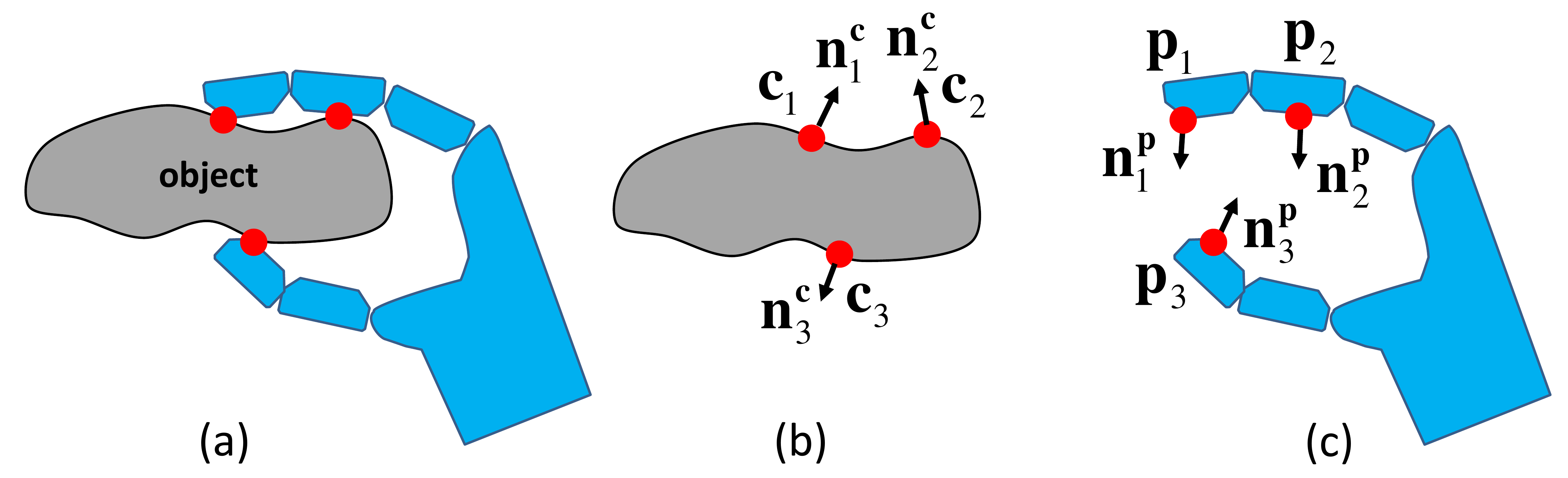,width=0.45\textwidth, page=1}
\caption{Grasp configuration and contacts. (a) A grasp configuration. (b) Contacts on the object surface. (c) The pre-defined points on the hand.}
\label{fig:notation}
\vspace{-1em}
\end{figure}

\begin{figure}[htb]
\centering
\epsfig{file=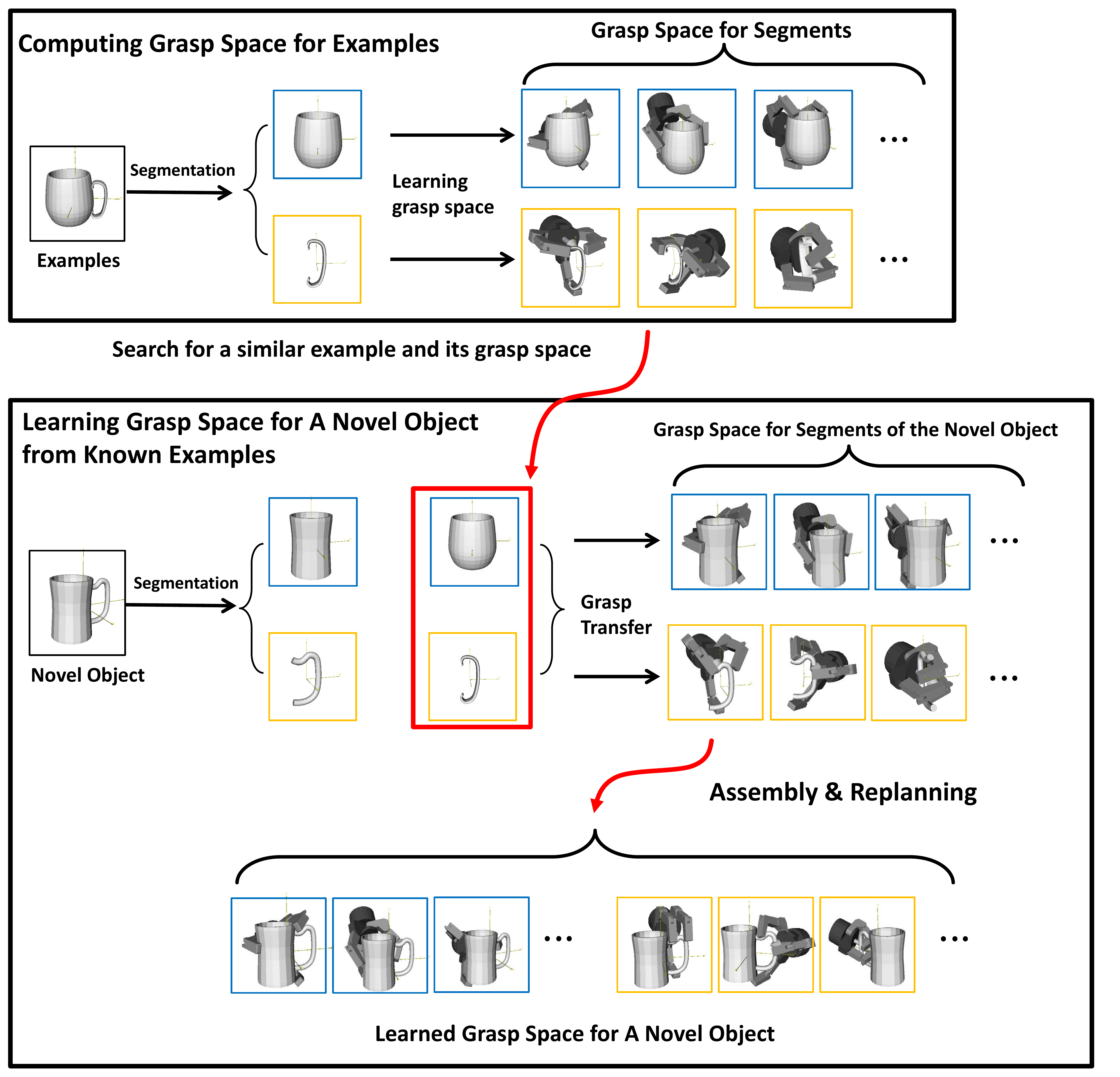,width=0.5\textwidth, page=1}
\caption{Transferring grasp configurations from a prior example to a novel object using active learning and local replanning.
Top: Computing stable grasp space for a given prior example using active learning. Bottom:
Grasp transfer to a novel object using bijective contact mapping and local replanning. The local replanning step not only generates a grasp that is similar to the origin, but it can also ensure the new grasp is stable.
} 
\label{fig:overview}
\vspace{-1em}
\end{figure}
Assume that a multi-fingered hand consists of $k$ joints and the joint variables are $\Theta=\{\theta_1,\theta_2,\cdots,\theta_k\}$.
Due to the high number of finger joints, the grasp space is a high-dimensional space, which is a subset of the configuration space constructed using a multi-fingered hand and an object to be grasped. A grasp space corresponds to a set of stable grasp configurations at which an object can be firmly grasped.
To grasp an object, a multi-fingered robotic hand must have multiple contacts with the object. As shown in Figure~\ref{fig:notation}, a grasp configuration corresponds to a few contacts between the hand (orange) and the object (grey). The contacts on the object surface is denoted by $\cc$ and its normal are denoted by $\nci$. The predefined points on the hand is denoted by $\pp$ and its normal is denoted by $\npi$.

\subsection{Algorithm Overview}
Figure~\ref{fig:overview} gives an overview of our algorithm. For an example object to be grasped, we first perform semantic segmentation and then compute the grasp space for each segment part.
We use SVM-based active learning to compute an initial approximation of the configuration space for each segment part. Active learning allows efficiently searching high dimensional configuration space.
We approximate the grasp space using particle swarm optimization. We use a new hybrid grasp measure to determine stable grasp configurations, taking account of both hand-object shape fitting and grasp quality defined in wrench space. In order to transfer the grasp space of the example object to other novel objects, 
we build bijective contact mapping between the corresponding parts of the example object and the novel object. Through contact points mapped to the novel object, its feasible grasp configuration is determined. Finally, we assemble individual parts of the novel object and use local replanning to adjust grasp configurations to ensure its stability and physical constraints.

\section{Approximating Configuration Space using Active Learning}\label{sec:learning}
Given an example object to be grasped, we first segment it into semantic parts. Next, we compute an approximation of the grasp space for each part using active learning. We randomly sample the configuration space and compute the collision state for each sample configuration using a discrete collision detection algorithm. There are two possible collision states: in-collision or collision-free, which correspond to a scenario in which the robotic hand collides with the object and a scenario in which it doesn't, respectively. Given a set of samples, we use an SVM technique and active learning iteratively to train a binary classifier to approximate the configuration space and the resulting decision boundary separates all in-collision configurations from the collision-free configurations.

\begin{figure}[htb]
\centering
\epsfig{file=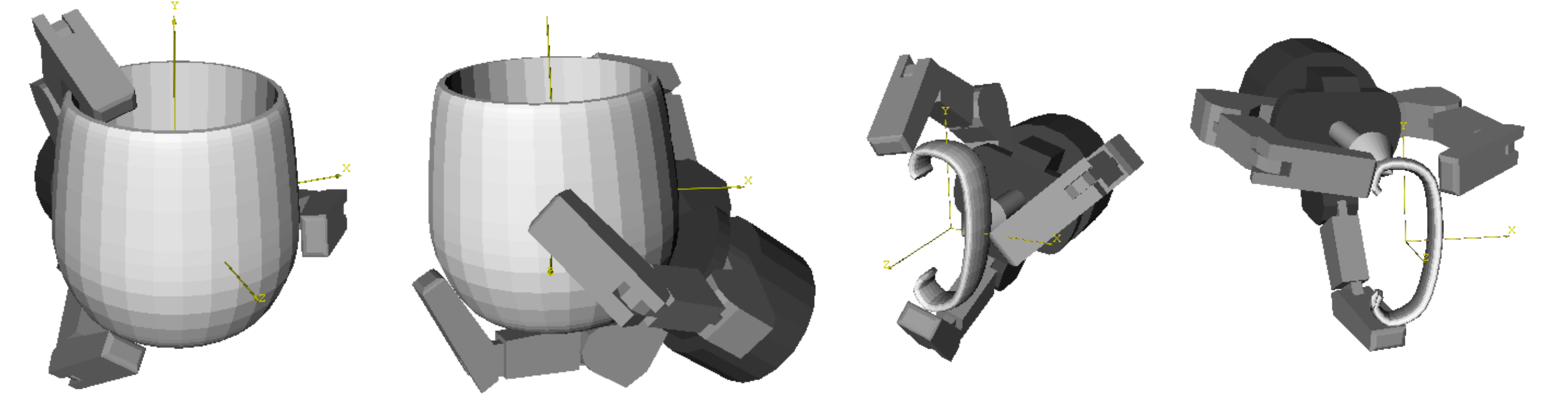,width=0.45\textwidth,page=1}
\caption{Learning grasp space for a known example object. Our algorithm is based on SVM-based active learning and particle swarm optimization.}
\label{fig:graspspace}
\vspace{-1em}
\end{figure}

Our goal is to compute a grasp space for each individual part of the prior example object. The discovery of a stable grasp configuration starting from a random configuration in the configuration space can be formulated as an optimization problem.
We define the following new \emph{hybrid grasp measure} to model the shape fitting and grasping stability between a hand posture and an object.  This new measure takes account of both hand-object shape fitting and  grasp quality defined in wrench space. Suppose we have a set of pre-specified points on the hand~\cite{ciocarlie2007dexterous}, the hybrid grasp measure is formulated as
\begin{align}\label{eq:matching1}
\sum_{i}\sum_{j}{(\omega_{1}|\pp_i-\cc_j| + \omega_{2}(1 - \npi \cdot \frac{\pp_i-\cc_j}{|\pp_i-\cc_j|}))} + 
 \omega_{3}\lg\frac{1}{\epsilon},
\end{align}
where the first term $|\pp_i-\cc_j|$ is the distance between the $i$th pre-defined point $\pp_i$ on the hand and its corresponding closest point $\cc_j$ on the object surface. The second term is based on the angle between the surface normal $\npi$ at $\pp_i$ and $\pp_i-\cc_j$. The last term is related to grasp quality~\cite{ferrari1992planning} defined in wrench space. $\epsilon\in(0,1]$ when the grasp is stable. A small $\epsilon$ indicates a relatively small external disturbance that can break a grasp's stability. $\omega_{1}$, $\omega_{2}$, and $\omega_{3}$ are the weights for the three terms, respectively.

We minimize the hybrid grasp measure with respect to the hand pose and joint variables $\Theta$ using {particle swarm optimization}~\cite{Clerc2002swarm} and then determine the stable grasps on the example object.
This formulation is highly non-linear~\cite{Pokorny2013qualityevaluation} and small changes in either hand position or finger postures can drastically alter the quality of the resulting grasp.
The stochastic nature of particle swarm optimization makes it a particularly good choice for solving the problem. Since a new configuration used in particle swarm optimization is generated as a  neighbor of the current configuration, we can avoid collisions during the sampling process. In addition, the computation of a gradient is unnecessary and therefore the particle swarm optimization algorithm is particularly applicable to non-linear functions in Equation~\ref{eq:matching1}.
We treat collision-free support vectors generated from the learning stage as the initial particles with an measurement value.
Given random searching velocities at the beginning of the optimization algorithm, the particles are updated in an iterative manner based on their velocities.
The particle swarm optimization algorithm records the global best position among all the particles and the local best position for each particle. The particle's movement is influenced by its local best position and the global best position. As a result, every particle is expected to move towards its best position, w.r.t. the given hybrid grasp measure, and the sparsely distributed particles can fully explore configuration space.
When particles explore the configuration space, the hand may collide with the object or have self-collisions between fingers. We use continuous collision detection~\cite{redon2005fast} to compute the first instance of contact that avoids collisions.

When a feasible grasp configuration is obtained, the force closure  can be determined using the contacts between the fingers and the object.
The grasp quality is then computed, which is related to the third term in the hybrid grasp measure. If a grasp configuration is stable, we keep it in the grasp space.
As a result, each grasp space is approximated by a set of discrete stable grasp configurations.
For an example object consisting of semantic parts, we compute the grasp spaces (i.e. a set of discrete stable grasp configurations) for all the individual parts.

\section{Transferring Grasp Space}\label{sec:Transferring}
In this section, we first present our bijective contact mapping for transferring grasp contact points from an example object to a novel object. Then we use grasp transfer and local replanning to obtain feasible and stable grasps for each part of the novel object.

\subsection{Bijective Contact Mapping}
Given a segmented part of an example object and a segmented part of a novel object, two sets of 3D points are uniformly sampled from the surfaces of the two objects. Let $\A=\{\ap:\ap\in\Tspace\}$ be a set of points on the example object and let $\B=\{\bp:\bp\in\Tspace\}$ be a set of points on the novel object. Assuming the two sets have the same number of points, the goal of bijective contact mapping is to find the correspondences between $\A$ and $\B$.

First, we compute a rigid alignment, a transformation from $\A$ to $\B$, to match the corresponding parts of the two objects. The resulting rigid alignment will be able to tolerate shape deviations between the example shape and the novel shape. The transformation $(\RR\in SO(3),\TT\in \Tspace)$ between the two objects is computed by minimizing the deviations:
\begin{align*}
\mathop{\arg\min}_{(\RR^{*},\TT^{*})}\sum_{\ap\in \A}\|\RR\ap+\TT-\ba\|^{2} + \arccos^{2}(\frac{\nba}{\|\nba\|}\cdot \RR\frac{\na}{\|\na\|}),
\end{align*}
where $\ba$ is denoted as the nearest neighbor point of $\ap$ in $\B$. $\na$ and $\nba$ are the normal of $\ap$ and $\ba$, respectively. Using this formulation, we obtain the transformation of the rigid alignment $\RR^{*}$ and $\TT^{*}$.

Second, we determine a bijective contact mapping between the two sets of points $\A$ and $\B$.
Since the rigid alignment between $\A$ and $\B$ has guarantee the corresponding points very close to one another,
we further refine these correspondences using local surface details such as point proximity and normal vectors. In addition, we use both forward contact mapping and backward contact mapping, defined as follows.

{\emph{Forward Mapping}:} We define a subset of points $\Bd = \{\bp\in\B|\nb\cdot\RR^{*}\na>\cos\delta \}$. Any point in $\Bd$ can find the corresponding point in $\A$ and the two points have at most an angle bias $\delta$ between their normal vectors.
The forward mapping of point $\ap\in\A$ is defined as
\begin{align}\label{eq:forward}
\F(\ap) = {\arg\min}_{\bp\in\Bd} \|\RR^{*}\ap+\TT^{*}-\bp \|.
\end{align}
\noindent{\emph{Backward Mapping}:} Analogously, we define a subset of points $\Ad = \{\ap\in\A|\na\cdot\RR^{*}\nb>\cos\delta \}$ and define the backward mapping of point $\bp\in\B$
\begin{align}\label{eq:backward}
\Back(\bp) = {\arg\min}_{\ap\in\Ad} \|\RR^{*}\ap+\TT^{*}-\bp \|.
\end{align}
We set $\delta=\frac{1}{6}\pi$ for the angle bias between two normal vectors.  Using both the forward and backward mappings, we can determine a mapping between the two sets of points $\A$ and $\B$.

Third, as suggested in ~\cite{hillenbrand2010non}, we compute a mapping from the domain of $\A$ to the domain of $\B$ using the interpolation of the point correspondences.
For any point $\xp$ on the example object, we search $\A$ for a subset of nearest points. We denote the subset as $\Ax$. The forward map of this point $\xp$ on the novel object is computed using the average of the forward maps of the closest points $\Ax$. The forward mapping of $\xp$ is
\begin{align}\label{eq:forwardMapping}
{\F}(\xp) = \frac{1}{\|\Ax\|}\sum_{\ap\in\Ax}\textsl{F}(\ap),
\end{align}
where $\|\Ax\|$ is the number of elements in $\Ax$. Analogously, we compute a backward mapping for a point $\yp$ as
\begin{align}\label{eq:backwardMapping}
{\Back}(\yp) = \frac{1}{\|\By\|}\sum_{\bp\in\By}\textsl{F}^{-1}(\bp),
\end{align}
where $\By\subset\B$ and it is the set of nearest neighbors of $\yp$.

\begin{figure}[htb]
\centering
\epsfig{file=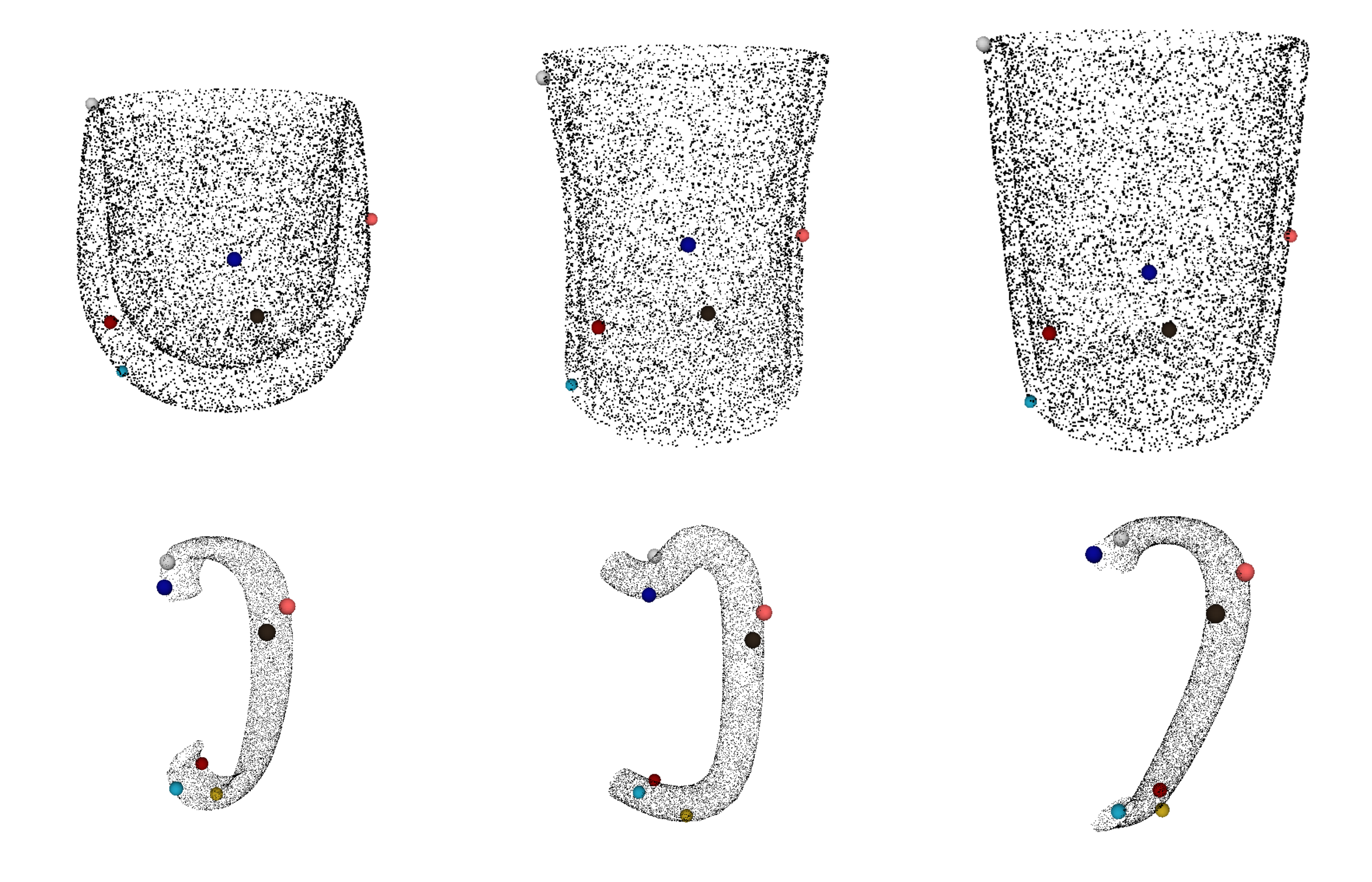,width=0.30\textwidth, page =1}
\caption{Bijective Contacts Mapping. The contacts on the given example objects (left column) are mapped to the novel objects (right two columns).}
\label{fig:warping}
\vspace{-1em}
\end{figure}

Finally, we use a forward-backward consistency check of these mappings to ensure the symmetry of the interpolated correspondences. For a point $\xp$ on the example object, its forward-backward mapping is $\FF^{-1}(\FF(\xp))$. We use the following condition to perform forward-backward consistency checking: $\|\xp-\FF^{-1}(\FF(\xp))\|<\gamma$,
where $\gamma$ is a user-specified tolerance and is related to the object dimension. If both the forward and backward mappings pass the consistency check, we can use the forward mapping to determine the points on the novel object. Figure~\ref{fig:warping} shows some examples of bijective contact mapping.

\subsection{Local Replanning for Feasible Grasps}\label{sec:LocalReplanning}

When directly applying the grasp configuration and the contacts to a novel object, it is very likely that there are slight penetrations between the multi-fingered hand and the novel object or that the grasp is not physically stable. Therefore, we use a local replanning technique to generate a stable grasp.
We first adjust the joint variables of the hand.
Let a finger's contact position be $\cc$ and the point on the novel object be $\dw$. We use the following conditions to generate a new grasp: $\cc_i =\dw_i$ and $\nci \cdot \ndwi = -1$,
where $\nci$ and $\ndwi$ are the normal vectors of contact points $\cc_i$ and $\dw_i$, respectively. The first condition ensures that the pair of corresponding contact points (i.e. the one on the fingers and the other on the object) match in terms of the position. The second condition ensures they have an opposite normal direction.
In addition, we limit the range of the joint variable to avoid large finger motions: $\theta_{i}^{low}\leq\theta_{i}\leq\theta_{i}^{up}$,
where $\theta_{i}^{low}$ and $\theta_{i}^{up}$ are the lower and upper bounds of the $i$th joint $\theta_{i}$, respectively. Based on ~\cite{borst2002calculating}, we define a new objective function subject to the above constraints. By incorporating contact points, normals, joint angles, and especially force closure-based grasp
quality, we obtain a grasp by solving for the joint variables $\bigTheta^{\ast}$.
\begin{align}\label{eq:partObjectiveFunc}
\mathop{\arg\min}_{\bigTheta^{\ast}}\sum_{i}(\mu_1\|\cc_i(\bigTheta)-\dw_i\|^2+ e^{\mu_2(\nci(\bigTheta)\cdot \ndwi)})+\notag \\
         \sum_{k}(e^{\mu_3(-\theta_{k}+\theta_{k}^{low})}+e^{\mu_3(\theta_{k}-\theta_{k}^{up})})+ \mu_4\lg\frac{1}{\epsilon},
\end{align}
where $\mu_1$, $\mu_2$, $\mu_3$ and $\mu_4$ are the weights for contact points, normals, joint angles, and grasp quality, respectively.
The first term accounts for the distance between finger contacts $\cc_i(\bigTheta)$ and contacts $\dw_i$ on the novel object. The second term computes their normal bias. The third term limits a joint's movement within a given interval. The last term is relevant to grasp quality $\epsilon$. If the objective function converges and yields a set of joint variables $\Theta^{\ast}$, a stable grasp is obtained for the novel object.


\begin{figure}[htb]
\centering
\subfigure[]{\epsfig{file=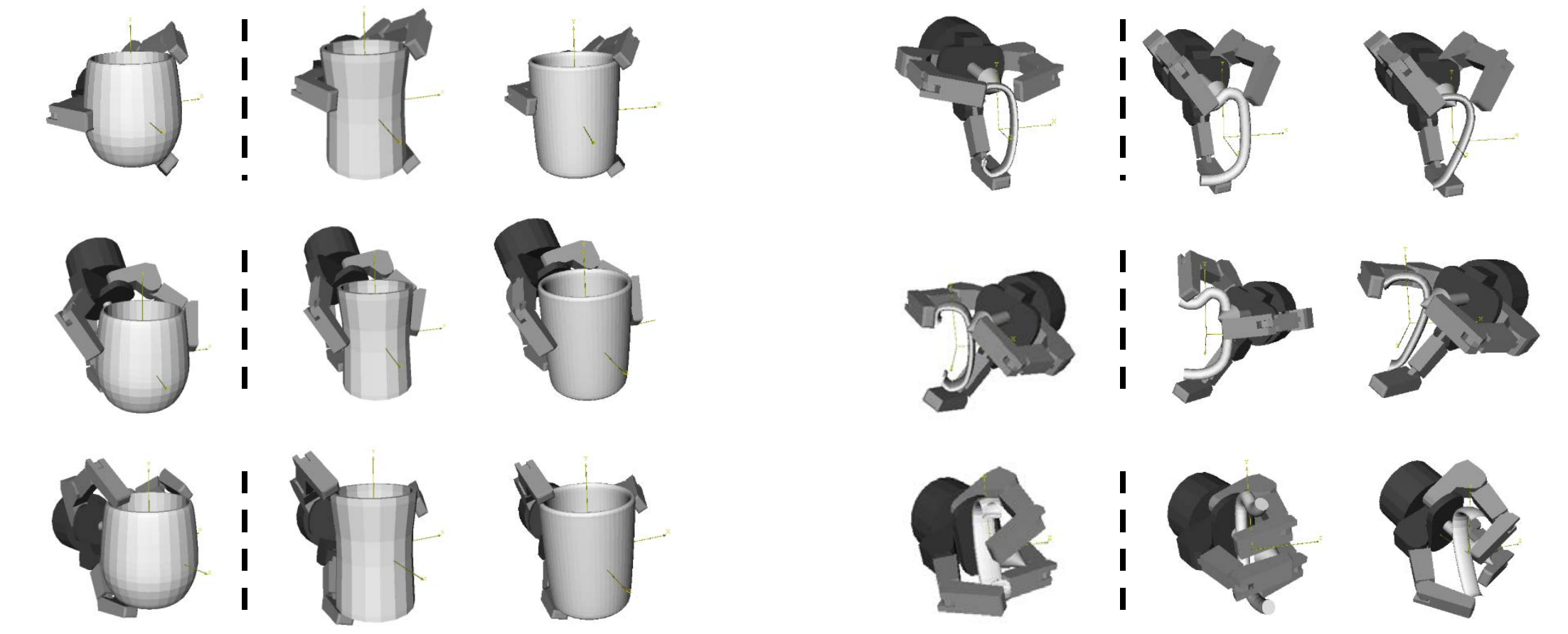,width=0.5\textwidth, page =1}}\\
\subfigure[]{\epsfig{file=./PartTransferResult.pdf,width=0.5\textwidth, page =1}}
\caption{Generating stable grasps using local replanning. The left three columns and the right three columns show results of different parts. The column on the left side of the dashed line shows stable grasps for each part of known example objects. The two columns on the right side of the dashed line are the resulting transferred grasps for novel objects.
}
\label{fig:part-transfer1}
\vspace{-1em}
\end{figure}
To solve Equation~\ref{eq:partObjectiveFunc}, we use a simulated annealing algorithm~\cite{kirkpatrick1983optimization} to explore the configuration space and search for a stable grasp.
The simulated annealing algorithm starts from the best grasp configuration that is initialized with the grasp learned for the example object. In each iteration, we sample the configuration space around the best grasp configuration and check the collision state for each sample. If the sample is in-collision, we explore the configuration space and add more samples (e.g., adding 20 samples at a time).
If the new samples are still in-collision, we discard it and proceed to the next iteration. If the sample is collision-free and has a better function value than the current best configuration, we replace the best configuration with this sample. The process proceeds in an iterative manner to find a best configuration. For any collision-free sample, we close the fingers, hold the object, and compute its grasping quality using the algorithm described in~\cite{ferrari1992planning}. If it is stable, we store it as a candidate grasp and repeat the iteration.
Figure~\ref{fig:part-transfer1} shows a few results after grasp transfer and local replanning. 

\subsection{Grasp Assembling and Replanning}\label{sec:partgrasp}

\begin{figure}[htb]
\centering
\subfigure[]{\epsfig{file=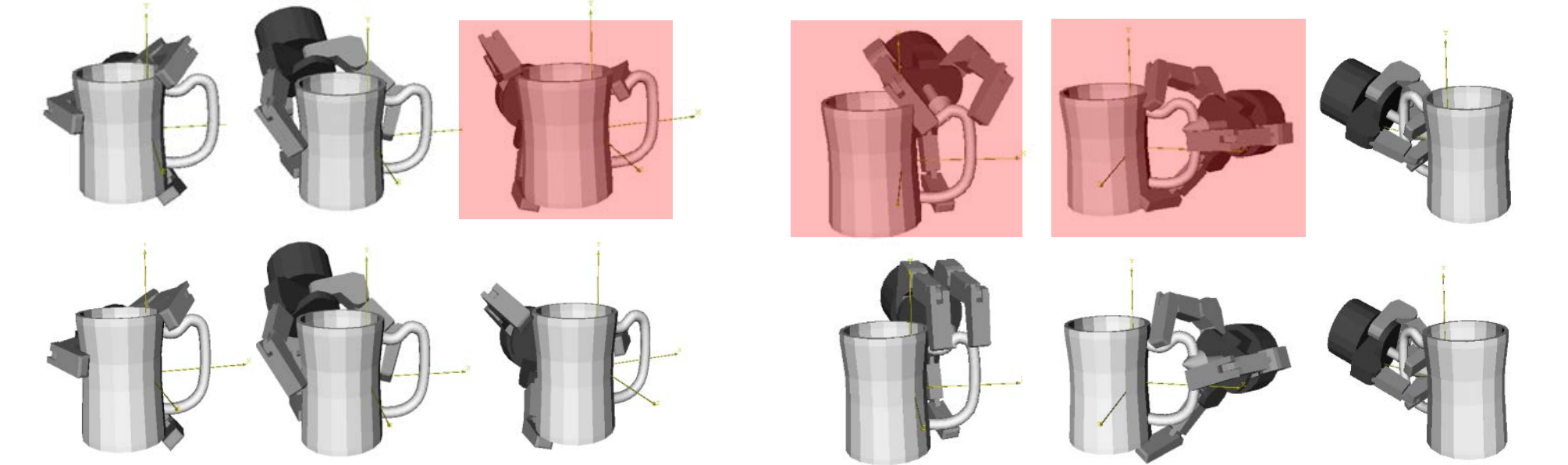,width=0.45\textwidth, page =1}}\\
\subfigure[]{\epsfig{file=./PartIntegrationResult.pdf,width=0.45\textwidth, page =2}}
\caption{Grasp configurations of novel objects before and after part assembly and local replanning. The upper row shows the assembly grasps without replanning; some grasps (highlighted in red) cause collisions with other object parts. We use local planning to compute a feasible and stable grasp, as shown in the bottom row.
}
\label{fig:assembly1}
\vspace{-1em}
\end{figure}

\begin{figure}[htb]
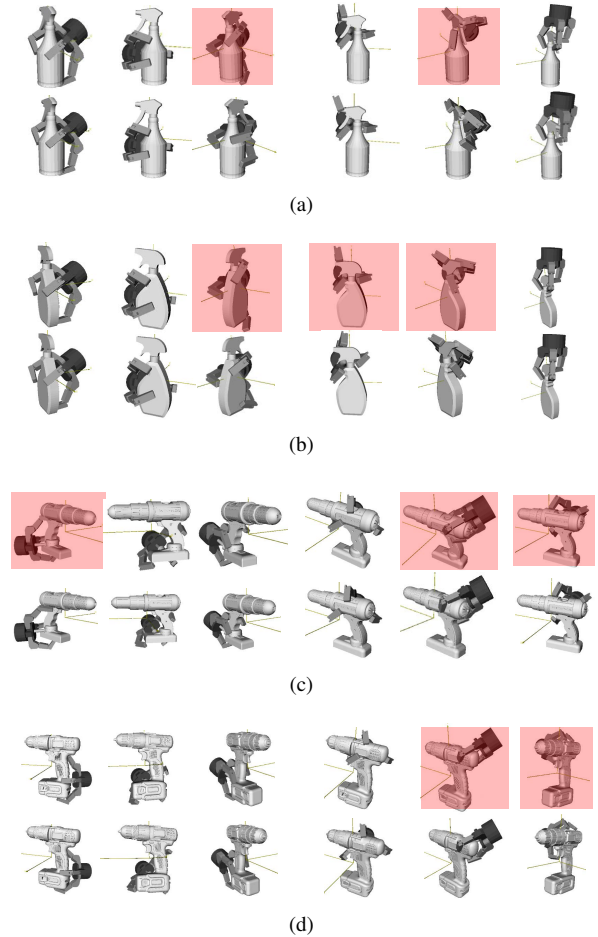

\centering
\subfigure[]{\epsfig{file=./PartIntegrationResult.pdf,width=0.45\textwidth, page =3}}\\
\subfigure[]{\epsfig{file=./PartIntegrationResult.pdf,width=0.45\textwidth, page =4}}\\
\subfigure[]{\epsfig{file=./PartIntegrationResult.pdf,width=0.45\textwidth, page =5}}\\
\subfigure[]{\epsfig{file=./PartIntegrationResult.pdf,width=0.45\textwidth, page =6}}\\
\caption{Grasp configurations for novel objects after part assembly and local replanning. Some grasps (highlighted in red) cause collisions with other object parts. Local replanning is able to find a feasible and stable grasp, as shown in the row below.
}
\label{fig:assembly2}
\end{figure}

Since the grasp configurations are computed for each part of the novel object, we need to assemble all the individual parts using their original semantic segmentation. We collect a set of grasps resulting from each part. However, these resulting grasps may not be stable for the assembled object or might result in collisions between the fingers and the object. Therefore, we examine whether a grasp causes any collisions between the hand and the novel object. If any collisions occur between the palm of the hand and the object, we discard the grasp. Note that it is non-trivial to adjust the position of the palm to generate a stable grasp since a slight adjustment may cause significant changes to the fingers. If a grasp results in collisions between a finger and the object, or if a grasp is collision-free but unstable, we use the replanning algorithm introduced in Section~\ref{sec:LocalReplanning} to make adjustments and to generate a stable grasp configuration.
Figure~\ref{fig:assembly1} and Figure~\ref{fig:assembly2} show some grasp configurations before and after part assembly and local replanning. 

\section{Experimental Results}\label{sec:results}
In this section, we present our benchmark models, some implementation details,
and the performance.

\subsection{Benchmark Objects}
\begin{figure}[htb]
\centering
\subfigure[example mug, $\sharp$1 novel object (mug), $\sharp$2 novel object (mug)]{\epsfig{file=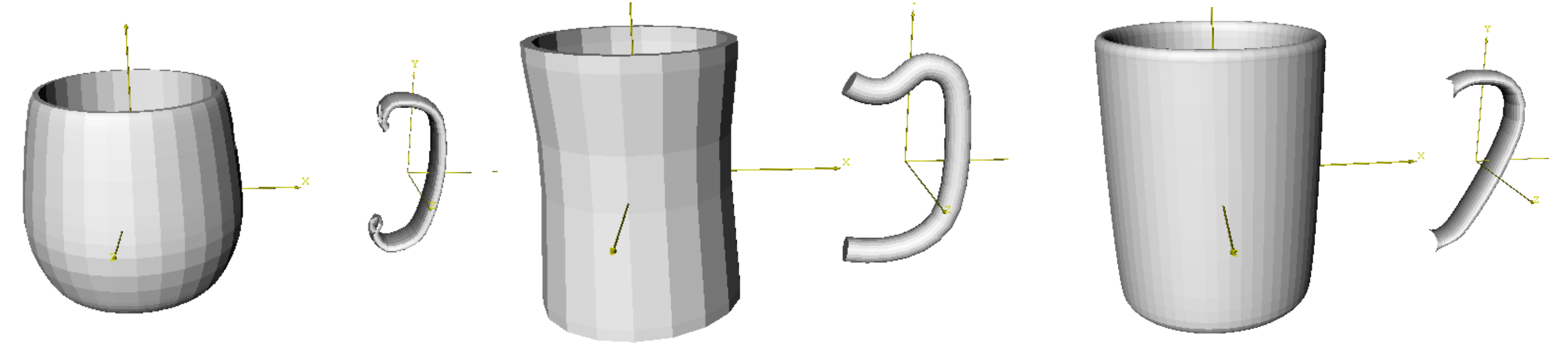,width=0.45\textwidth, page =1}}\\
\subfigure[example spray bottle, $\sharp$1 novel object (spray bottle), $\sharp$2 novel object (spray bottle)]{\epsfig{file=./obj-cat.pdf,width=0.45\textwidth, page =2}}\\
\subfigure[example power drill, $\sharp$1 novel object (power drill), $\sharp$2 novel object (power drill)]{\epsfig{file=./obj-cat.pdf,width=0.45\textwidth, page =3}}
\caption{Our experiments use three categories of objects and their segmentations. (a) mug; (b) spray bottle; (c) power drill. The objects in the first column are known examples and the others are novel objects.
}
\label{fig:obj-cat}
\end{figure}

We evaluated the performance of our algorithm on three categories of objects using a wide variety of grasps.
The complexity of objects also shown in Table~\ref{tab:learninggraspspace} and Table~\ref{tab:transfertime}
We used a three-fingered Barrett hand to grasp objects.
The categories of the benchmark objects include mug, spray bottle, and power drill.
In our experiments, we chose one object in each category as an example object and the other two as novel objects.
We compute grasp configurations for the segmented parts of each example object and then transfer the grasps to the novel objects.
As shown in Figure~\ref{fig:obj-cat}, the leftmost object is selected as the example object.

\begin{figure}[htb]
\centering\epsfig{file=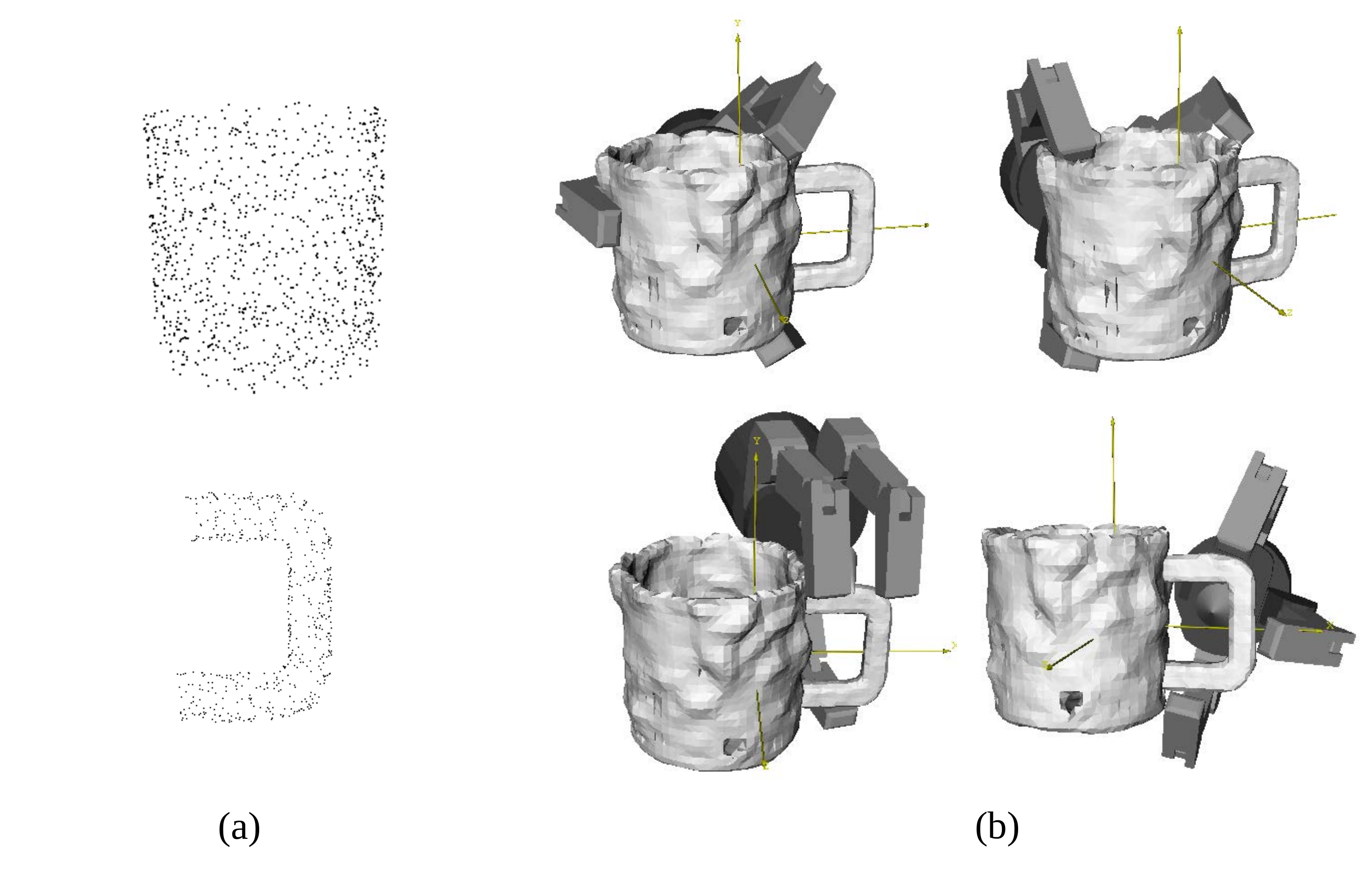,width=0.42\textwidth, page =1}
\caption{The result grasps transferred from known example object for reconstructed models. (a) point clouds; (b) grasp results.
}
\label{fig:pointcloudgrasp-mug}
\vspace{-1em}
\end{figure}

In robot mapping and navigation, depth cameras are commonly used for the capture of the objects and the surrounding environment. Then the reconstruction of 3D point cloud models are used to perform robotic grasping task. In our experiment, we also apply our algorithm to the objects reconstructed from noisy point cloud, as shown in Figure~\ref{fig:pointcloudgrasp-mug}.


\subsection{Implementation}
We implement our transfer algorithm using GraspIt!~\cite{miller2004graspit}. We use a shape diameter function (SDF)~\cite{shapira2008consistent} to segment the given objects into parts.
When computing the grasp space for each segmented part of the example object, we set $\omega_{1} = 0.02$ and $\omega_{2} = 1.0$, as suggested in~\cite{Ciocarlie2010Low}. In addition, we set $\omega_{3} = 20$. We sampled $40,000$ training data as the initial input of SVM-based active learning to compute grasp spaces for all the example objects. We chose collision-free support vectors in SVM as particles during the particle swarm optimization process. The time require to learn the configuration space, the time to compute the grasp space and the number of stable grasps are shown in Table~\ref{tab:learninggraspspace}.
During grasp transfer, we set $\mu_1=10$, $\mu_2=5$, $\mu_3=5$, and $\mu_4=20$ in Equation~\ref{eq:partObjectiveFunc}. Table~\ref{tab:transfertime} shows the average time of transferring a stable grasp from an example object to the novel object.
\begin{table*}[!t]
  \centering
  \small
    \begin{tabular}{|c|c|c|c|c|}
    \hline
    example objects &  number of triangles & \tabincell{c}{learning contact space \\(seconds)} & \tabincell{c}{computing stable grasps\\ (seconds)} & $\#$ stable grasps\\
    \hline
     mug body & 1864 & 1343.6 & 1405.5 & 133\\
    \hline
     mug handle & 1586 & 1254.8 & 1287.1 & 60\\
    \hline
     spray bottle body & 1844 & 1404.5 & 1536.4 & 194\\
    \hline
     spray bottle head & 894 & 1286.7 & 1328.0 & 48\\
    \hline
     power drill body & 4893 & 1654.2 & 1604.2 & 61\\
    \hline
     power drill head & 5052 & 1795.1 & 1635.9 & 56\\
    \hline
    \end{tabular}
    \caption{Statistics of Learning Process: the model complexity, the time of active learning/computing stable grasps and the number of stable grasps. Our algorithm based on active learning can automatically find sufficient stable grasps in high-dimensional configuration space.}
  \label{tab:learninggraspspace}
\end{table*}

\subsection{Performance}
We measure grasp quality and grasp transfer success rates to evaluate the performance of our algorithm.

\begin{figure*}[htb]
\centering
\subfigure[$\sharp$1 mug body]{
\epsfig{file=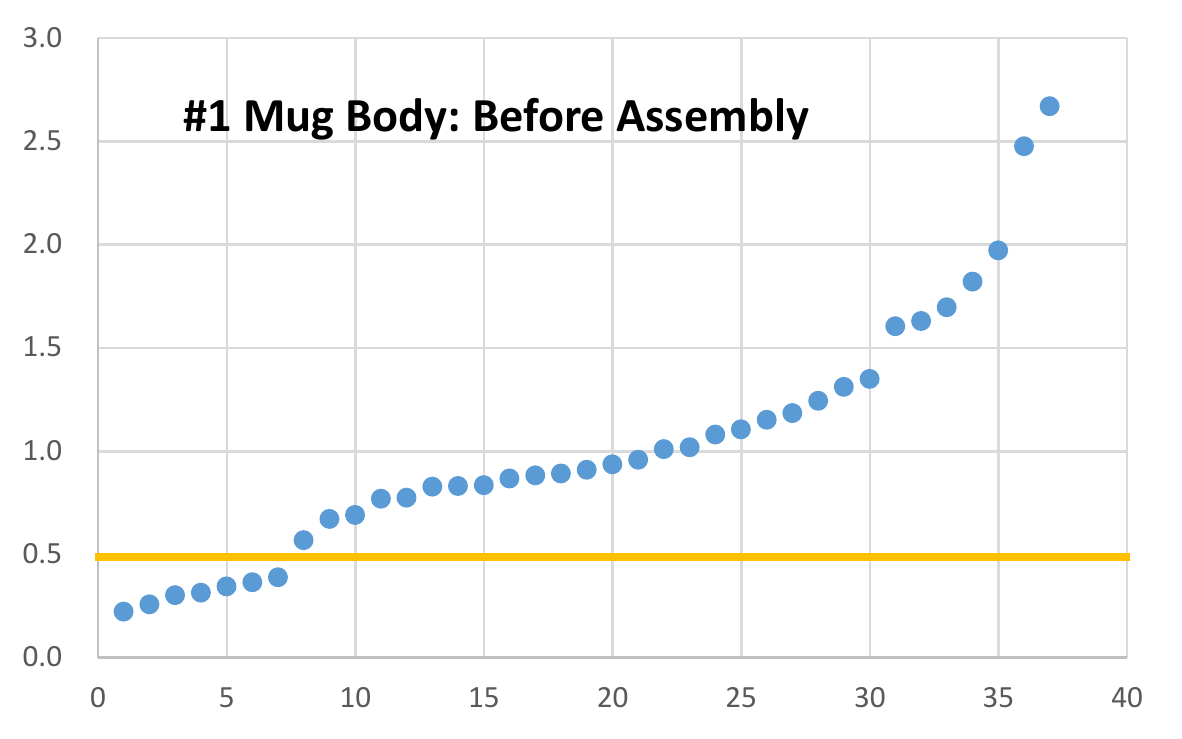,width=0.4\textwidth,page =1}
\epsfig{file=./quality.pdf,width=0.4\textwidth,page =2}
}\\
\subfigure[$\sharp$1 mug handle]{
\epsfig{file=./quality.pdf,width=0.4\textwidth, page =3}
\epsfig{file=./quality.pdf,width=0.4\textwidth, page =4}
}\\
\subfigure[$\sharp$2 mug body]{
\epsfig{file=./quality.pdf,width=0.4\textwidth,page =5}
\epsfig{file=./quality.pdf,width=0.4\textwidth,page =6}
}\\
\subfigure[$\sharp$2 mug handle]{
\epsfig{file=./quality.pdf,width=0.4\textwidth, page =7}
\epsfig{file=./quality.pdf,width=0.4\textwidth, page =8}
}\\
\caption{Grasp Quality before and after part assembly and replanning. horizontal axis is the grasp index and vertical axis is the grasp quality.
}
\label{fig:quality-mug}
\end{figure*}

\begin{figure*}[htb]
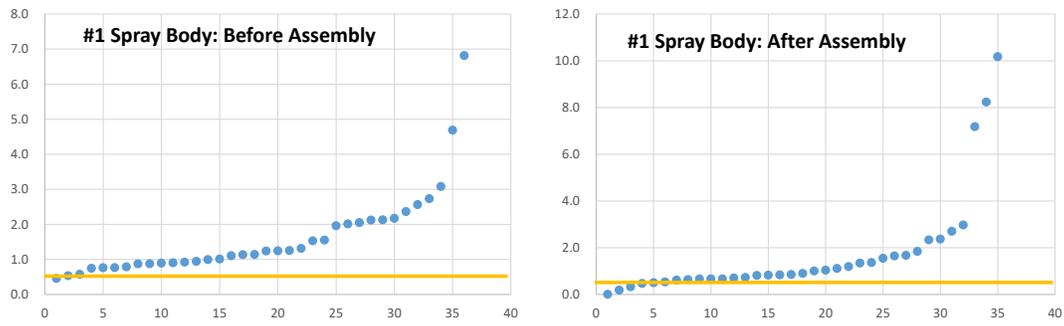
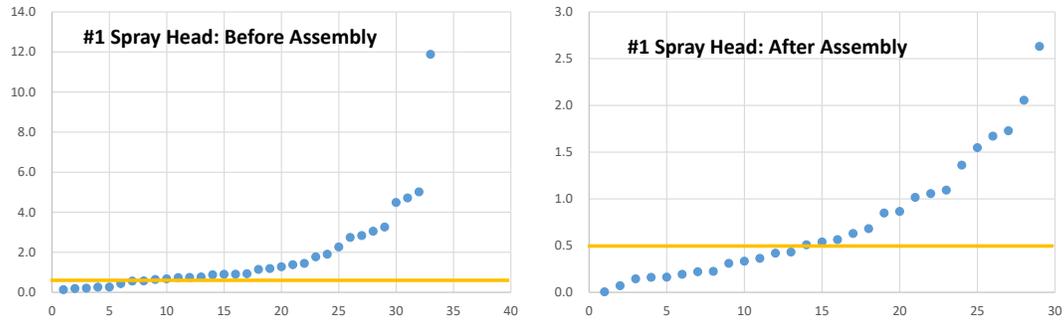
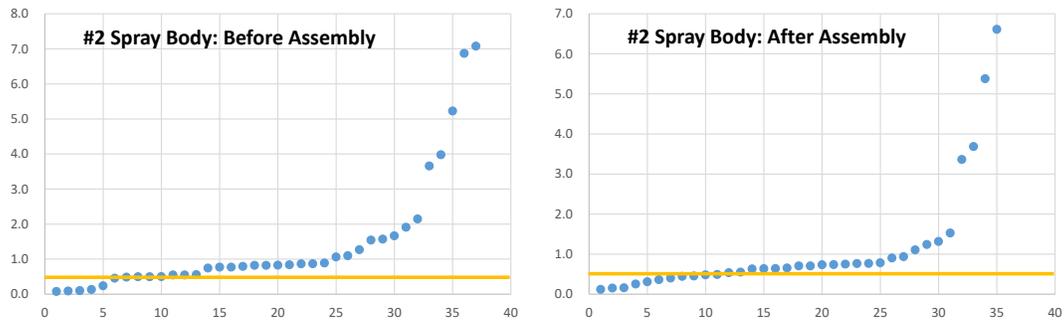
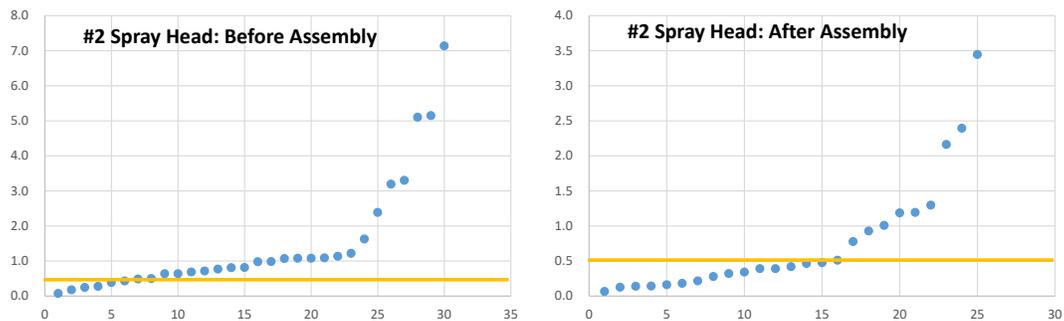

\centering
\subfigure[$\sharp$1 spray bottle body]{\epsfig{file=./quality.pdf,width=0.4\textwidth, page =9}
\epsfig{file=./quality.pdf,width=0.4\textwidth, page =10}}\\
\subfigure[$\sharp$1 spray bottle head]{\epsfig{file=./quality.pdf,width=0.4\textwidth, page =11}
\epsfig{file=./quality.pdf,width=0.4\textwidth, page =12}}\\
\subfigure[$\sharp$2 spray bottle body]{\epsfig{file=./quality.pdf,width=0.4\textwidth, page =13}
\epsfig{file=./quality.pdf,width=0.4\textwidth, page =14}}\\
\subfigure[$\sharp$2 spray bottle head]{\epsfig{file=./quality.pdf,width=0.4\textwidth, page =15}
\epsfig{file=./quality.pdf,width=0.4\textwidth, page =16}}
\caption{The ratio of the grasp quality for novel objects against example objects before part assembly and after part assembly and replanning. (a)(b) spray bottle model $\sharp$1; (c)(d) spray bottle model $\sharp$2.}
\label{fig:qualitystatistic-spray}
\end{figure*}

\begin{figure*}[htb]
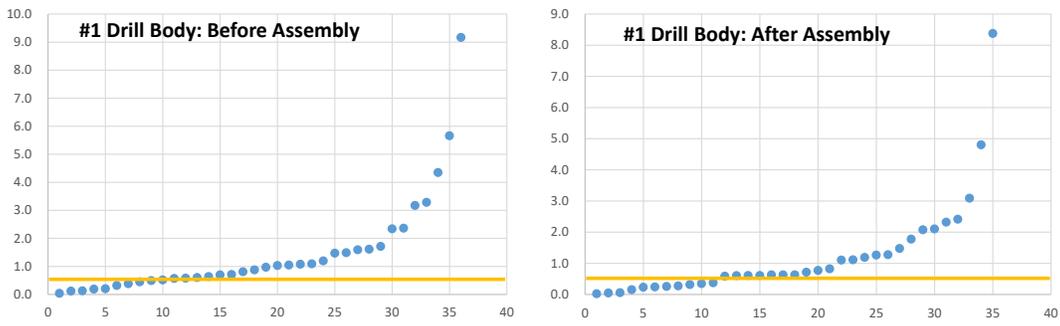
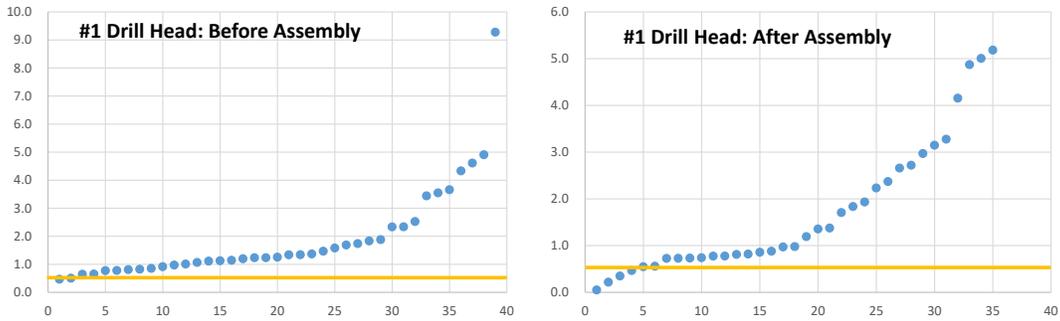
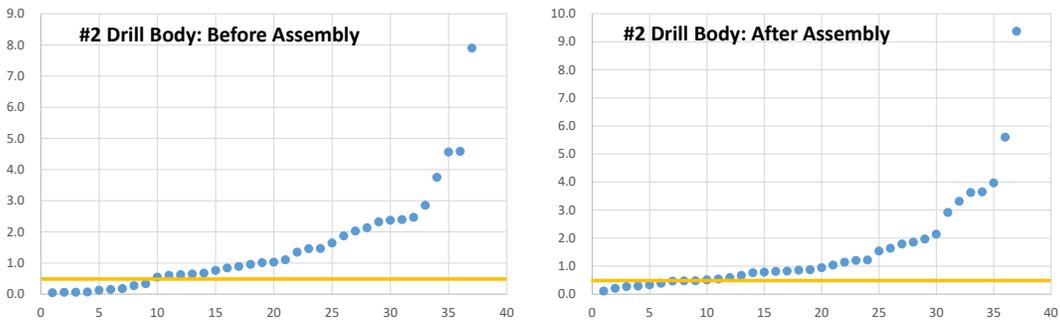
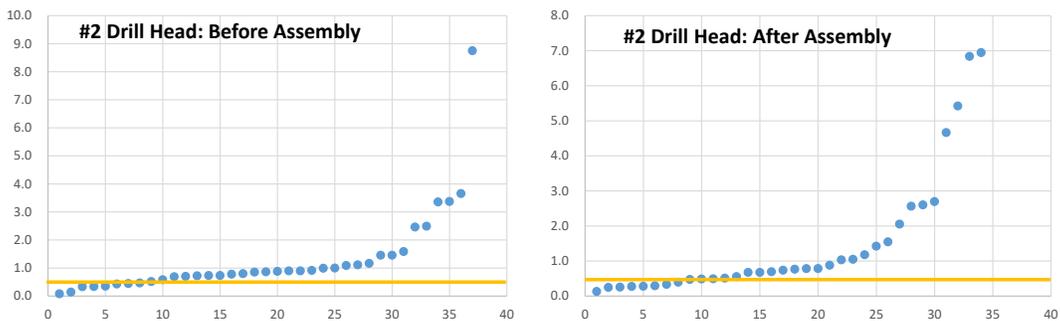

\centering
\subfigure[$\sharp$1 power drill body]{\epsfig{file=./quality.pdf,width=0.4\textwidth, page =17}
\epsfig{file=./quality.pdf,width=0.4\textwidth, page =18}}\\
\subfigure[$\sharp$1 power drill head]{\epsfig{file=./quality.pdf,width=0.4\textwidth, page =19}
\epsfig{file=./quality.pdf,width=0.4\textwidth, page =20}}\\
\subfigure[$\sharp$2 power drill body]{\epsfig{file=./quality.pdf,width=0.4\textwidth, page =21}
\epsfig{file=./quality.pdf,width=0.4\textwidth, page =22}}\\
\subfigure[$\sharp$2 power drill head]{\epsfig{file=./quality.pdf,width=0.4\textwidth, page =23}
\epsfig{file=./quality.pdf,width=0.4\textwidth, page =24}}
\caption{The ratio of the grasp quality for novel objects against example objects before part assembly and after part assembly and replanning. (a)(b) power drill model $\sharp$1; (c)(d) power drill model $\sharp$2.}
\label{fig:qualitystatistic-drill}
\end{figure*}

\begin{figure*}[htb]
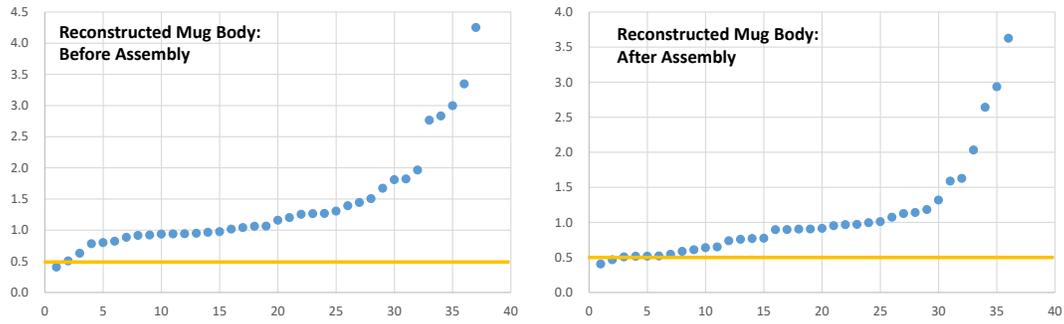
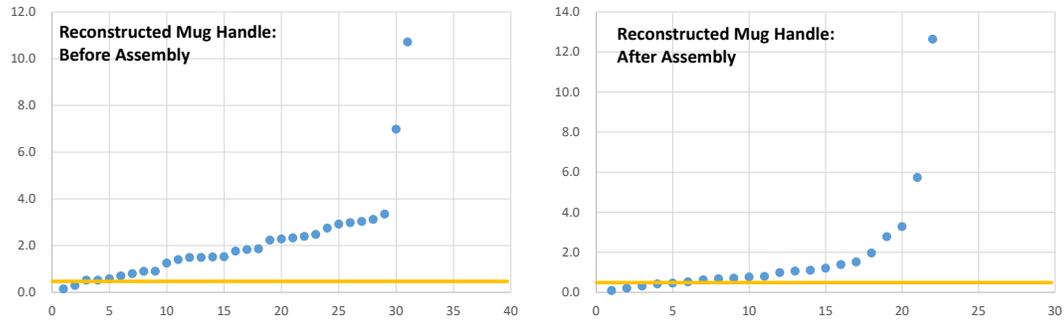

\centering
\subfigure[reconstructed mug body]{\epsfig{file=./quality.pdf,width=0.4\textwidth,page =25}
\epsfig{file=./quality.pdf,width=0.4\textwidth,page =26}}\\
\subfigure[reconstructed mug handle]{\epsfig{file=./quality.pdf,width=0.4\textwidth,page =27}
\epsfig{file=./quality.pdf,width=0.4\textwidth,page =28}}\\
\caption{The ratio of the grasp quality for model parts reconstructed from point cloud against example objects before part assembly and after part assembly and replanning. 
}
\label{fig:pointcloud-quality-mug}
\end{figure*}

{\bf Grasp Quality:}
We examine the grasp quality for novel objects against grasp quality for example objects.
As shown in Figures~\ref{fig:quality-mug}, ~\ref{fig:qualitystatistic-spray},~\ref{fig:qualitystatistic-drill}, and~\ref{fig:pointcloud-quality-mug}. For the mug body (see Figure~\ref{fig:quality-mug}), we found most of the resulting grasps are stable and $43.2\%$ grasps are more stable than the original grasps of example objects. For the mug handle, $45.5\%$ grasps are more stable than the original grasps.

\begin{figure*}[htb]
\centering
\subfigure[$\sharp$1 mug body]{
\epsfig{file=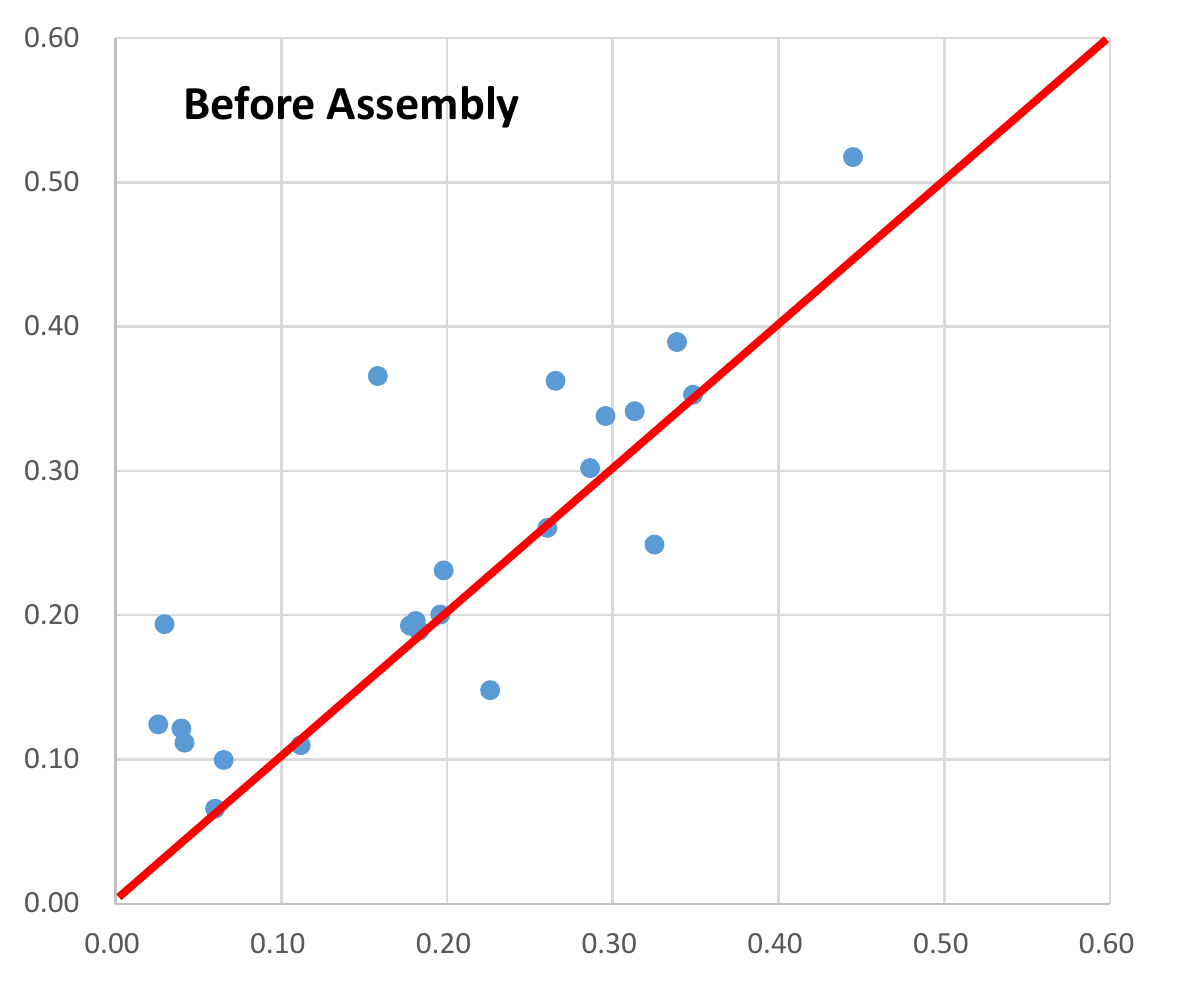,width=0.3\textwidth,page =1}
\epsfig{file=./compare.pdf,width=0.3\textwidth,page =2}
}\\
\subfigure[$\sharp$1 mug handle]{
\epsfig{file=./compare.pdf,width=0.3\textwidth, page =3}
\epsfig{file=./compare.pdf,width=0.3\textwidth, page =4}
}\\
\subfigure[$\sharp$1 mug body]{
\epsfig{file=./compare.pdf,width=0.3\textwidth,page =5}
\epsfig{file=./compare.pdf,width=0.3\textwidth,page =6}
}\\
\subfigure[$\sharp$1 mug handle]{
\epsfig{file=./compare.pdf,width=0.3\textwidth, page =7}
\epsfig{file=./compare.pdf,width=0.3\textwidth, page =8}
}\\
\caption{Grasp Quality Comparison. If the ratio is greater than 1 (i.e., above the red diagonal), our grasp (vertical axis) is better than the straightforward grasp (horizontal axis).
}
\label{fig:qualitycompare-mug}
\end{figure*}

\begin{figure*}[htb]
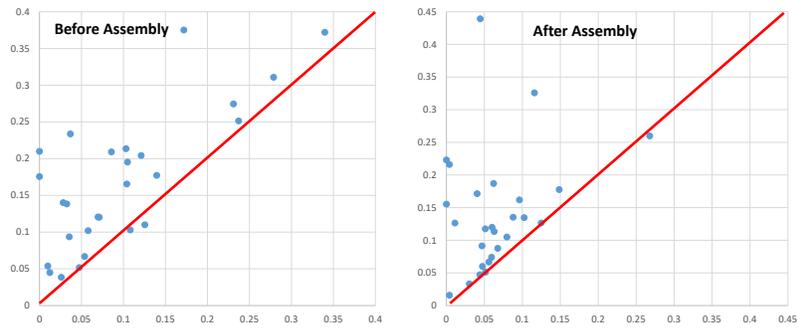
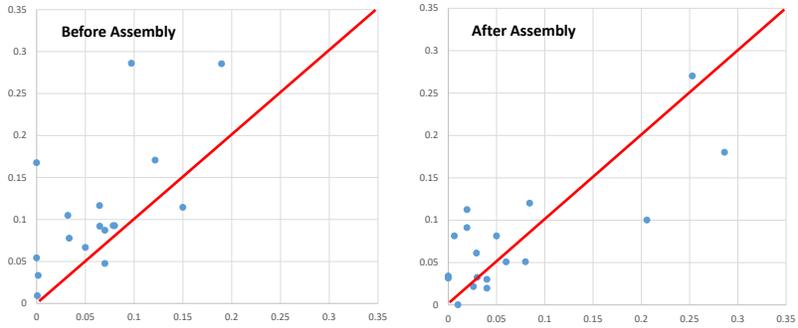
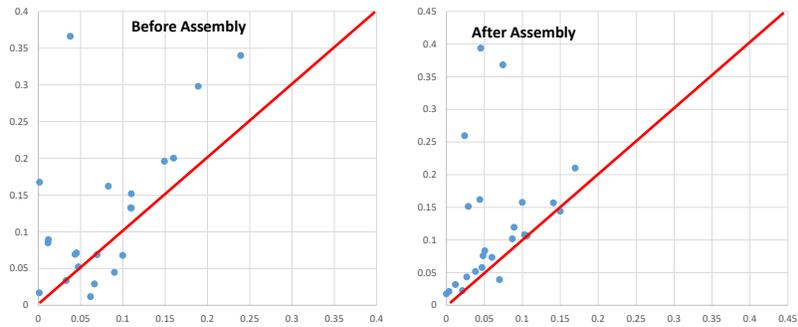
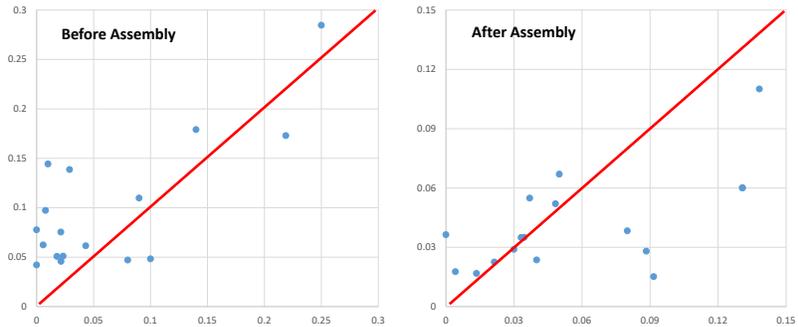

\centering
\subfigure[$\sharp$1 spray bottle body]{\epsfig{file=./compare.pdf,width=0.3\textwidth,page =9}
\epsfig{file=./compare.pdf,width=0.3\textwidth,page =10}}\\
\subfigure[$\sharp$1 spray bottle head]{\epsfig{file=./compare.pdf,width=0.3\textwidth, page =11}
\epsfig{file=./compare.pdf,width=0.3\textwidth, page =12}}\\
\subfigure[$\sharp$2 spray bottle body]{\epsfig{file=./compare.pdf,width=0.3\textwidth, page =13}
\epsfig{file=./compare.pdf,width=0.3\textwidth, page =14}}\\
\subfigure[$\sharp$2 spray bottle head]{\epsfig{file=./compare.pdf,width=0.3\textwidth, page =15}
\epsfig{file=./compare.pdf,width=0.3\textwidth, page =16}}
\caption{Grasp quality comparison between our method and a straightforward method (straightening the colliding fingers and enclose the object). The points distribute above the diagonal mean our method outperform the straightforward method. (a)(b) spray bottle model $\sharp$1; (c)(d) spray bottle model $\sharp$2.}
\label{fig:qualitycompare-spray}
\end{figure*}

\begin{figure*}[htb]
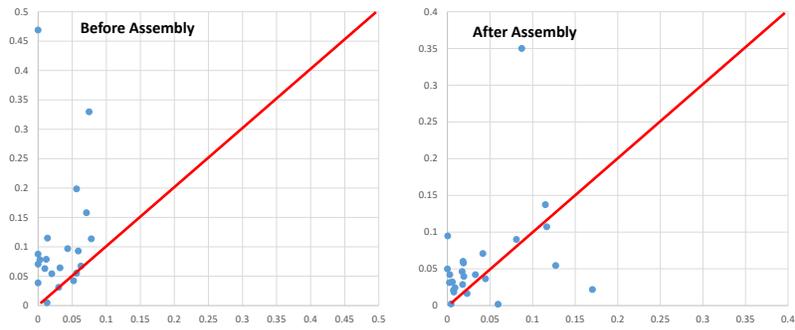
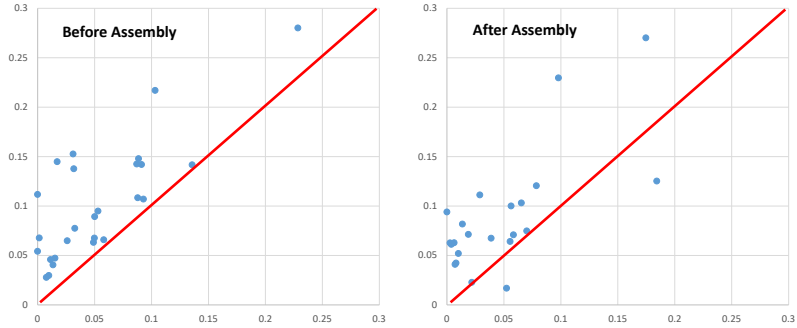
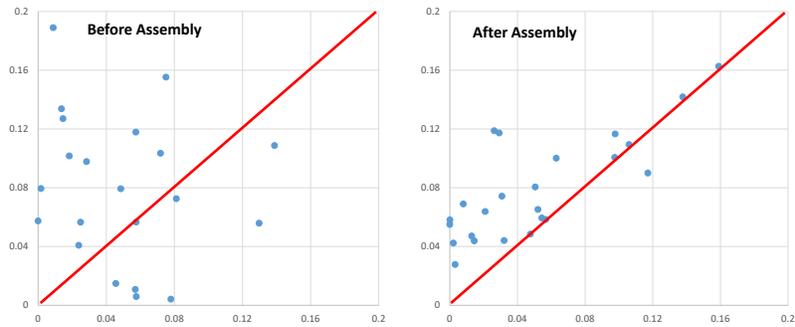
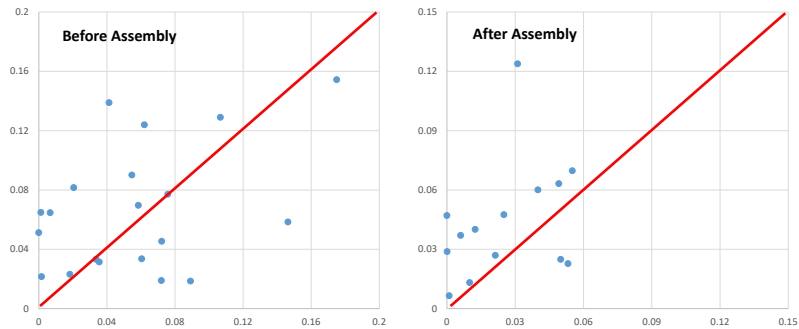

\centering
\subfigure[$\sharp$1 power drill body]{\epsfig{file=./compare.pdf,width=0.3\textwidth,page =17}
\epsfig{file=./compare.pdf,width=0.3\textwidth,page =18}}\\
\subfigure[$\sharp$1 power drill head]{\epsfig{file=./compare.pdf,width=0.3\textwidth, page =19}
\epsfig{file=./compare.pdf,width=0.3\textwidth, page =20}}\\
\subfigure[$\sharp$2 power drill body]{\epsfig{file=./compare.pdf,width=0.3\textwidth, page =21}
\epsfig{file=./compare.pdf,width=0.3\textwidth, page =22}}\\
\subfigure[$\sharp$2 power drill head]{\epsfig{file=./compare.pdf,width=0.3\textwidth, page =23}
\epsfig{file=./compare.pdf,width=0.3\textwidth, page =24}}
\caption{Grasp quality comparison between our method and a straightforward method (straightening the colliding fingers and enclose the object). The points distribute above the diagonal mean our method outperform the straightforward method. (a)(b) power drill model $\sharp$1; (c)(d) power drill model $\sharp$2.}
\label{fig:qualitycompare-drill}
\end{figure*}

\begin{figure*}[htb]
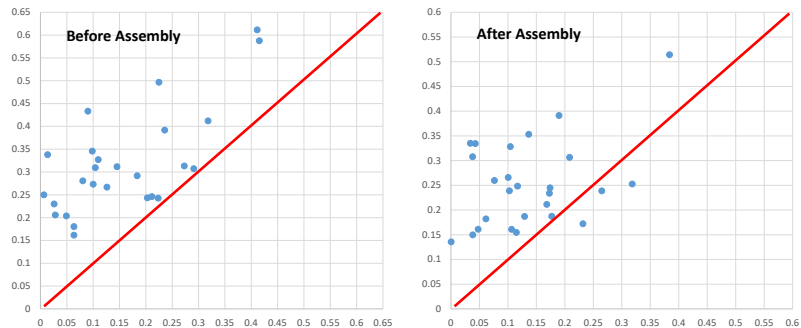
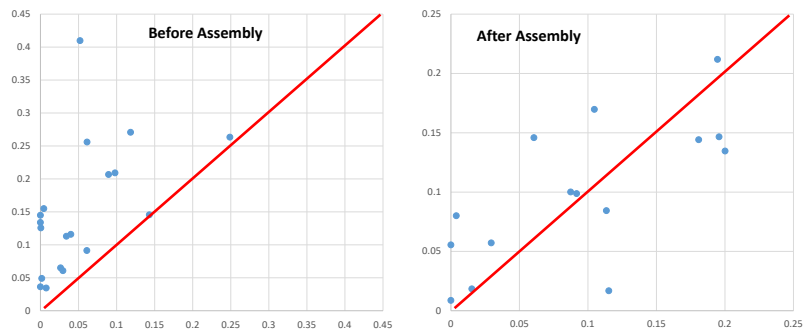

\centering
\subfigure[reconstructed mug body]{\epsfig{file=./compare.pdf,width=0.3\textwidth,page =25}
\epsfig{file=./compare.pdf,width=0.3\textwidth,page =26}}\\
\subfigure[reconstructed mug handle]{\epsfig{file=./compare.pdf,width=0.3\textwidth, page =27}
\epsfig{file=./compare.pdf,width=0.3\textwidth, page =28}}
\caption{Grasp quality comparison between our method and a straightforward method (straightening the colliding fingers and enclosing the object). The points distribute above the diagonal mean our method outperform the straightforward method.
}
\label{fig:pointcloud-qualitycompare-mug}
\end{figure*}

We compare our method with a prior grasp method based on open-close action~\cite{krug2014optimization}. The latter first straightens the colliding fingers and
close the fingers until they come into contact with the object. It does not use local replanning. As shown in Figures~\ref{fig:qualitycompare-mug},~\ref{fig:qualitycompare-spray},~\ref{fig:qualitycompare-drill}, and~\ref{fig:pointcloud-qualitycompare-mug}, our grasp is better than this naive open-close grasp if the ratio of their grasp quality is greater than 1 (i.e., above the diagonal), where
$77.2\%\sim84.8\%$ points distribute above the diagonal (i.e., our method outperforms the prior grasp method~\cite{krug2014optimization}).
{\bf Grasp Transfer Success Rate:}
We measure the grasp transfer success rate for the two stages: grasp transfer and part/grasp assembly, as shown in Table~\ref{tab:transfertime} and Table~\ref{tab:pointcloud-successrate}. Here, we define a grasp transfer is success if a grasp of the example object is successfully transferred to the novel object and the resulting grasp is eventually stable.
We compare our method with the prior method based on open-close action. As shown in Table~\ref{tab:transfertime} and Table~\ref{tab:pointcloud-successrate}, our method exhibits high success ratios and outperforms the prior method. Table~\ref{tab:transfertime} and Table~\ref{tab:pointcloud-successrate} list the number of successful grasp transfers and grasp assemblies. The first number is the result of our method and the second is the result of the prior method. We evaluate 40 stable grasps of the example object for each test.
Our method has a high success rate of transferring grasp configurations to novel objects and the success rate ranges from $72.5\%$ to $92.5\%$ in our experiment.

\subsection{Comparisons}
In this section, we compare our algorithm with state of the art methods~\cite{li2007data,vahrenkamp2016part,rodriguez2018transferring-skill} and highlight the benefits.
Li et al.~\cite{li2007data} proposed a data-driven approach using shape matching to grasp synthesis. They captured human grasps and transferred candidate grasps to a new object by matching the hand shape to the whole object shape via identifying collections of features.
Our method does not require pre-recorded human data. Moreover, our active learning algorithm can generate feasible and stable grasps for arbitrary objects, including high genus and complex topology. Our approach can be regarded as a complementary technique to task-specific robotic grasping. Vahrenkamp et al.~\cite{vahrenkamp2016part} presented a grasp planning approach that is capable of generating grasps that are applicable to familiar objects. Their approach is also based on semantic segmentation and grasp transfer to a novel object. In contrast, our method focuses on dexterous grasping planning for individual object parts. We use bijective contact mapping to transfer contacts and use local replanning to ensure the feasibility and stability of the new grasp. Rodriguez et al.~\cite{rodriguez2018transferring-skill} transferred manipulations skills to novel instances using a novel latent space non-rigid registration. They built the latent space for a category of objects in learning stage. With the latent space, an inference can be performed to find a transformation from the canonical model to the novel models. Grasps can be transferred to the novel model using this transformation. Our method is very general in the sense that it can deal with various models and be applied to high-DOF dexterous hands.

\begin{table*}[!t]
  \centering
  \small
    \begin{tabular}{|c|c|c|c|c|c|}
    \hline
    \multirow{2}{*}{novel objects} & \multirow{2}{*}{tris} & \multicolumn{2}{c|}{average time} & \multicolumn{2}{c|}{success rate}\\
    \cline{3-6}
     &  & transfer stage (s) & assembly stage (s) & transfer stage & assembly stage\\
    \hline
    $\#$1 mug body & 752 & 322.5&	305.3 & 37/35 & 35/29\\
    \hline
    $\#$1 mug handle & 1284 & 294.7&	302.5 & 33/21 & 29/19\\
    \hline
    $\#$2 mug body & 2550 & 335.3&	289.2 & 36/35 & 36/33 \\
    \hline
    $\#$2 mug handle & 1586 &283.1&	295.8 & 33/23 & 26/29 \\
    \hline
    $\#$1 spray bottle body & 2294 & 329.0&  293.4 & 36/29 & 35/29\\
    \hline
    $\#$1 spray bottle head & 1241 & 298.9&	345.6 & 32/17 & 22/14\\
    \hline
    $\#$2 spray bottle body & 2074 & 334.4& 315.1 & 34/22 & 32/19\\
    \hline
    $\#$2 spray bottle head & 737 &312.8&	364.2 & 28/15 & 23/18\\
    \hline
    $\#$1 power drill body & 6105 &414.7&	475.6 & 35/17 & 32/25\\
    \hline
    $\#$1 power drill head &6850& 437.5&	498.1 & 37/24 & 33/26\\
    \hline
    $\#$2 power drill body &6914& 421.5&	480.5 & 39/22 & 36/26\\
    \hline
    $\#$2 power drill head &6781& 453.4&	509.4 & 35/20 & 31/13\\
    \hline
    \end{tabular}
    \caption{Grasp Transfer Success Rate. Our method has up to $52.6\%$ improvement in success rate against straightforward method.}
  \label{tab:transfertime}
\end{table*}

\begin{table*}[!t]
  \centering
  \small
    \begin{tabular}{|c|c|c|c|}
    \hline
    \multirow{2}{*}{novel objects} & \multirow{2}{*}{number of points} & \multicolumn{2}{c|}{success rate}\\
    \cline{3-4}
    & & transfer stage & assembly stage \\
    \hline
    reconstructed mug body & 1000 & 38/34 & 37/31\\
    \hline
    reconstructed mug handle & 500 & 31/19 & 28/23\\
    \hline
    \end{tabular}
    \caption{The number of successful grasp transfers and grasp assemblies for the reconstructed models. We compare success rate of our method with straightforward method.}
  \label{tab:pointcloud-successrate}
\end{table*}

\subsection{Time Complexity}
Here, we analyze the time complexity of our algorithm.

\noindent{\bf Active Learning Configuration-Space:}
The time spent in the learning stage depends on the number of samples, the time of collision detection, and the time to learn an approximation of configuration space. The collision detection is accelerated using precomputed bounding volume hierarchies of finger links and objects.
The time spent stable grasp computation stage depends on the number of particles and the number of iterations. In each iteration, we perform collision detection for each hand configuration and compute its hybrid grasp measurement value.

\noindent{\bf Bijective Contact Mapping:}
The time spent on bijective contact mapping consists of three stages: rigid alignment computation, correspondence computation, and consistency checking. Its time spent depends on the number of iterations and the time for computing deviations between two objects in each iteration. The mapping computation and consistency checking depends on the number of points on the objects. We have to compute forward mapping for all points on the example object and backward mapping for all points on the novel object.

\noindent{\bf Replanning:}
The replanning is an optimization process, depending on the number of iterations and the time complexity of each iteration. In each iteration, we generate at most 20 samples. We perform collision detection and compute an objective function value for each feasible sample in configuration space.

\section{Conclusions, Limitations, and Future Work}\label{sec:conclusion}
In this paper,
we presented a new approach to grasping a novel object through learning the grasping knowledge of a known object. Our algorithm avoids computing or learning the grasp configurations from scratch. Our experiment shows that our algorithm works for a wide range of object categories.
{Our method has a high success rate of transferring grasp configurations to novel objects, ranging from $72.5\%$ to $92.5\%$ in our experiment. Our method has up to $52.6\%$ improvement in success rate against prior method.}

Our algorithm has some limitations. First, our algorithm assumes the object-space representation of a complex object is available. Second, it also relies on the object category classification and the quality of segmentation. Fortunately, many model databases~\cite{wohlkinger20123dnet,shapenet2015} have provided a large number of categories and segmented objects. In our future work, we are interested in extending our algorithm to handle partially-known novel models and very high-genus models. Moreover, we would like to improve the performance using hardware acceleration so that real-time performance can be achieved. 

\bibliographystyle{abbrv-doi}
\bibliography{template}

\end{document}